\documentclass[conference]{IEEEtran}
\IEEEoverridecommandlockouts
\usepackage{cite}
\usepackage{amsmath,amssymb,amsfonts}
\usepackage{algorithmic}
\usepackage{graphicx}
\usepackage{textcomp}
\usepackage{xcolor}
\usepackage{subfigure}
\usepackage{makecell}
\usepackage{multirow}

\usepackage[T1]{fontenc}
\usepackage[utf8]{inputenc}

\begin{document}
\bstctlcite{IEEEexample:BSTcontrol} 

\title{An Empirical Evaluation of the Risks of Artificial Intelligence Model Updates Using Clinical Data: Stability, Arbitrariness, and Fairness
\thanks{This work was funded by the Spanish Ministry of Science and Innovation (REF: DIN2021-011865).  It was also partially supported by the Department of Research and Universities of the Government of Catalonia (SGR 00930), the EU-funded project SoBigData++ (grant agreement 871042), and by MCIN/AEI /10.13039/501100011033 under the Maria de Maeztu Units of Excellence Programme (CEX2021-001195-M). None of the funders played any role in the study design, data collection, analysis, interpretation of results, or the writing of this manuscript.}
}


\author{\IEEEauthorblockN{Ioannis Bilionis}
\IEEEauthorblockA{\textit{Adhera Health and}\\
\textit{Universitat Pompeu Fabra}}
Barcelona, Spain \\
0000-0002-1435-4358
\and
\IEEEauthorblockN{Ricardo C. Berrios}
\IEEEauthorblockA{\textit{Adhera Health, USA}}
0009-0008-5449-0267
\and
\IEEEauthorblockN{Luis Fernandez-Luque}
\IEEEauthorblockA{\textit{Adhera Health, Spain}}
0000-0001-8165-9904
\and
\IEEEauthorblockN{Carlos Castillo}
\IEEEauthorblockA{\textit{Universitat Pompeu Fabra} \\
\textit{and ICREA}}
Catalonia, Spain \\
0000-0003-4544-0416}

\maketitle

\begin{abstract}
Artificial Intelligence (AI) and Machine Learning (ML) models used in clinical settings are increasingly deployed to support clinical decision-making. 
However, when training data become stale due to changes in demographics, environment, or patient behaviors, model performance can degrade substantially. While updating models with new training data is necessary, such updates may also introduce new risks.
We evaluated the proposed monitoring framework on four publicly available U.S.-based Type 1 Diabetes datasets containing high-resolution continuous glucose monitoring (CGM) data, comprising approximately 11,300 weekly observations from 496 participants younger than 20 years. 
All datasets included structured sociodemographic information. 
Using the prediction of severe hyperglycemia events in children with Type 1 Diabetes as a case study, we examine how different model update strategies can adversely affect model stability by causing predictions to change for a large number of cases after retraining, increase prediction arbitrariness, and worsen subgroup fairness and the balance of error rates across populations.
We propose multiple dimensions for continuous monitoring to detect these issues and argue that such monitoring is essential for the development of trustworthy clinical decision support systems.


\end{abstract}


\section{Introduction}
In recent years, artificial intelligence (AI) and machine learning (ML) models have become increasingly integrated into clinical workflows, supporting diagnostic, prognostic, and therapeutic decision-making across a wide range of medical applications.
Advances in biomedical sensing technologies and digital health infrastructures have enabled continuous patient data collection through new medical devices and monitoring systems.
At key consideration when using AI/ML models in clinical scenarios is that evolving disease phenotypes, together with shifts in environmental, nutritional, and behavioral factors at the population level, introduce systematic changes in data distributions that can degrade model performance over time~\cite{vela2022temporal,hatherley2025moving}.
As a result, model updating and maintenance have become critical requirements in clinical settings where data are generated continuously~\cite{feng2022clinical}.
Effective updating strategies must allow models to adapt to distributional drifts and newly observed clinical patterns, while mitigating bias~\cite{khoshravan2023impact}, catastrophic forgetting ~\cite{mermillod2013stability,van2024continual} and ensuring reliable, safe, equitable and trustworthy decision support for clinicians~\cite{meijerink2025updating}.

A growing body of work has investigated update strategies and methodological frameworks for continual learning in hospital and clinical environments.
Prior studies have proposed data-driven approaches for adapting predictive models to evolving data distributions~\cite{zhang2025data}, encompassing drift-aware retraining ~\cite{patchipala2023tackling,subasri2025detecting,casimiro2022towards}, training set selection~\cite{florence2025retrain,wei2023representative,shen2024data}, model optimization~\cite{dissanayake2024continuous}, incremental learning paradigms~\cite{leonard2025mitigating}, as well as system degradation detection methods ~\cite{guan2025keeping} and self-healing~\cite{rauba2024self} mechanisms during model updates and maintenance.
Despite these advances, relatively limited attention has been devoted to the interaction between continual learning strategies and algorithmic fairness~\cite{davis2022open,kim2024development,rahman2025adaptive,ceccon2025fairness}, particularly with respect to sociodemographic bias and subgroup-level performance stability over time in real-world clinical settings.

Beyond aggregate predictive performance, the deployment of AI/ML models in clinical settings raises critical concerns related to fairness, stability, and trustworthiness. 
Importantly, performance-preserving updates do not necessarily imply stable or robust decision behavior: as models are retrained, the multiplicity~\cite{ganesh2025curious} of equally performant yet behaviorally divergent models may induce shifts in prediction stability~\cite{bertsimas2024towards}, giving rise to arbitrariness.
Such phenomena can affect individual-level decisions without manifesting as significant changes in overall performance metrics, posing substantial risks in high-stakes clinical applications~\cite{ueda2024fairness,wang2023watch,naher2024measuring,bilionis2025disparate,paulus2020predictably}.

In this context, concepts such as model multiplicity~\cite{jain2025allocation,dai2025intentional,kobylinska2024exploration}, prediction stability~\cite{bertsimas2024towards}, and model  arbitrariness~\cite{cooper2024arbitrariness} become essential dimensions for model assessment alongside accuracy and algorithmic fairness metrics.
Furthermore, uncertainty-aware mechanisms~\cite{angelopoulos2021gentle}, including abstention strategies based on frameworks such as conformal prediction, offer a principled means to 
mitigate unreliable predictions during deployment and retraining.
Together, these considerations motivate a holistic evaluation of continual learning systems beyond conventional metrics~\cite{bansal2019updates}, integrating performance, fairness, stability, arbitrariness, multiplicity, and uncertainty management into model update and maintenance pipelines.
%

From a clinical and regulatory standpoint, deploying AI/ML models as Software as a Medical Device (SaMD)  poses challenges in validation, monitoring, and post-deployment updating, with limited guidance on systematic model updates for time-stamped clinical data under strict privacy constraints~\cite{feng2022clinical}.
To move beyond fixed models, regulatory agencies such as the U.S. Food and Drug Administration now support controlled model evolution such as the Predetermined Change Control Plan (PCCP), yet few approved AI medical devices report post-deployment retraining, revealing a gap between regulatory guidance and practice~\cite{wu2024regulating}.
%

In this paper, we study the behavior of clinical machine learning models under continual retraining, extending evaluation beyond predictive performance to include stability, arbitrariness, and algorithmic fairness. 
We analyze how these characteristics evolve across successive retraining phases and compare alternative update strategies in terms of their ability to balance accuracy, fairness, and robustness over time. 
To support safe model adaptation in dynamic clinical settings, we further examine uncertainty-aware abstention and propose a 
clinical decision support pipeline that abstains for out-of-distribution cases.
Our experimental framework provides interpretable, longitudinal insights into model behavior under distributional shift, advancing the governance of adaptive clinical AI systems with an emphasis on equity, robustness, and patient safety.



\section{Methods}
We employ a modular continual learning evaluation framework that simulates real-world clinical model updates while jointly assessing performance, fairness, temporal stability, and uncertainty-aware safety mechanisms.

\subsection{Continual Learning Experimental Design}
We consider a supervised clinical prediction task defined over patient-level longitudinal data. Let $\mathcal{D} = \bigcup_{t=0}^{T} \mathcal{D}_t$ denote the full dataset, where each batch $\mathcal{D}_t$ corresponds to a temporally ordered cohort (e.g., admission period or data acquisition window). Each batch contains samples $(x_i, y_i, a_i)$, where $x_i \in \mathbb{R}^d$ denotes features, $y_i \in \{0,1\}$ the clinical outcome, and $a_i \in \mathcal{A}$ a protected attribute (e.g, Sex).


In the retraining phase $t$, the model is trained on all past data
\begin{equation}
\mathcal{D}^{\text{train}}_t = \bigcup_{k=0}^{t-1} \mathcal{D}_k,
\label{eq:retrain_data}
\end{equation}
reflecting cumulative learning from previously observed patients.

We compare multiple retraining strategies to capture common deployment practices assessing trade-offs between performance, multiplicity, stability, and fairness:
\begin{itemize}
    \item \textbf{Full retraining}: the model is reinitialized and trained from scratch using $\mathcal{D}^{\text{train}}_t$.
    \item \textbf{Last-batch retraining}: retraining is performed using only the most recent batch $\mathcal{D}_{t-1}$.
    \item \textbf{Subset retraining}: retraining on a fixed-size subset of samples, randomly drawn from all previously observed batches, with the subsample size chosen to match the average batch size.
    \item \textbf{No retraining}: the initial model is deployed without updates.
\end{itemize}

We performed an initial benchmarking of multiple machine learning algorithms to establish baseline performance and assess model stability across datasets.
Four classification schemes—Logistic Regression, Random Forest, CatBoost, and Na\"ive Bayes were compared, reporting mean AUC and variability.
Two training strategies were evaluated: including the protected feature (“fairness through awareness”) and excluding it (“fairness through unawareness”). 
Across datasets, Logistic Regression was used as base classifier of the experiments since it consistently demonstrated the highest stability and overall performance. 



\begin{figure}[!b] 
    \centering
    \includegraphics[width=\columnwidth]{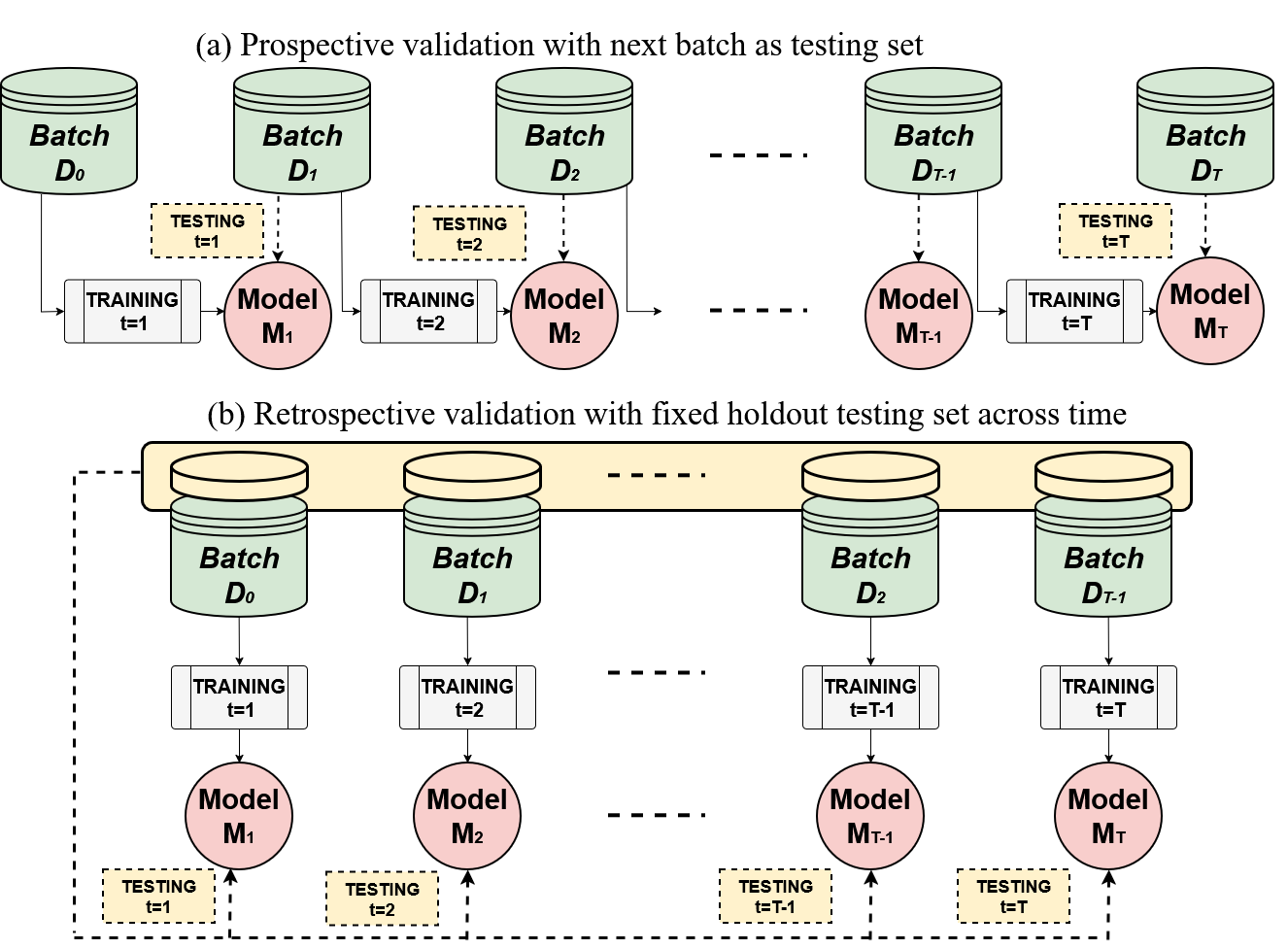}
    \caption{\textbf{Illustration of model evaluation schemes}. 
    (a) In the prospective setting, the model is evaluated on the immediately subsequent batch, reflecting a realistic deployment scenario where future data are unseen at training time. 
    (b) In the retrospective setting, a fixed stratified patient-level holdout subset is defined once and used for evaluation  across all retraining phases. 
    The illustration is for the last-batch retraining strategy, but the same evaluation schemas apply to the other retraining strategies (full cumulative retraining, random subsampling of past data, and no retraining), differing only in the composition of the training set used at each retraining phase.}
    \label{fig:evaluation_schemas}
\end{figure}

Furthermore, we adopt two complementary evaluation settings, depicted in Fig.~\ref{fig:evaluation_schemas}:
\begin{itemize}
    \item \textbf{Prospective evaluation}: the model trained at phase $t$ is evaluated on the immediately following batch $\mathcal{D}_t$, simulating what we would observe in a real-world deployment.
    \item \textbf{Retrospective fixed hold-out evaluation}: a stratified patient-level subset (10\%) is sampled as a fixed test set and evaluated across all retraining phases. This allows us to control for changes in task difficulty, and must be repeated $N$ times with different random seeds to ensure measure robustness.
\end{itemize}


\subsection{Evaluation Metrics}
\subsubsection{Predictive Performance}
Predictive accuracy and deviations from accuracy equity can be assessed using the area under the receiver operating characteristic curve (AUC).
%


At each retraining phase $t$, we compute the overall AUC as well as group-conditional AUCs for each protected subgroup $a \in \mathcal{A}$:
\begin{equation}
\mathrm{AUC}_t^{(a)} = \mathrm{AUC}\big( \{(x_i,y_i) \in \mathcal{D}_t : a_i = a\} \big)
\label{eq:group_auc}
\end{equation}
Performance evolution over time is quantified via temporal differences capturing gains or degradation induced by retraining:
\begin{equation}
\Delta \mathrm{AUC}_t = \mathrm{AUC}_t - \mathrm{AUC}_{t-1}
\label{eq:auc_diff_temporal}
\end{equation}
To explicitly measure performance disparity, we compute the absolute AUC gap between protected subgroups at each time moment, reflecting unequal predictive performance across subpopulations and enabling tracking trade-offs of algorithmic fairness and performance during retraining:
\begin{equation}
\mathrm{AUC}_{\text{diff},t} = \left| \mathrm{AUC}_t^{(a_1)} - \mathrm{AUC}_t^{(a_2)} \right|
\label{eq:auc_gap}
\end{equation}

\subsubsection{Fairness Metrics}
Algorithmic fairness is evaluated at each retraining phase with respect to a protected attribute $a \in \{a_1,a_2\}$. In addition to subgroup-specific AUCs, we quantify it using the following standard criteria.
Both metrics are computed independently at each retraining phase to monitor temporal fairness drift induced by continual model updates.

\textbf{Equal Opportunity (EO)} measures disparities in true positive rates (TPR) across protected groups:
\begin{equation}
\Delta \mathrm{EO} = \left| \mathrm{TPR}^{(a_1)} - \mathrm{TPR}^{(a_2)} \right|
\label{eq:fair_eo}
\end{equation}

\textbf{Demographic Parity (DP)} captures differences in positive prediction rates between groups:
\begin{equation}
\Delta \mathrm{DP} = \left| P(\hat{y}=1 \mid a=a_1) - P(\hat{y}=1 \mid a=a_2) \right|
\label{eq:fair_dp}
\end{equation}

\subsubsection{Stability and Prediction Consistency}

Beyond performance and fairness, we evaluate the stability of model predictions during retraining. Stability is assessed through a hierarchy of complementary metrics capturing global, temporal, and local sources of prediction variability.

\textbf{Overall Self-Consistency (SC)} quantifies the agreement of model predictions under repeated retraining and resampling. Following prior work on model arbitrariness~\cite{cooper2024arbitrariness}, SC is estimated via bootstrap resampling and reflects the probability that independently retrained models produce identical predictions for the same input.

\textbf{Systematic Arbitrariness (SA)} measures distributional divergence in self-consistency between protected groups, capturing structured instability that may disproportionately affect specific subpopulations.

\textbf{Temporal Self-Consistency} (TSC) summarizes the evolution of prediction stability across retraining phases. Let $\mathrm{SC}_i(t) \in [0.5,1]$ denote the bootstrap-based self-consistency of individual $i$ at retraining phase $t$. The temporal self-consistency of individual $i$ is defined as:
\begin{equation}
\mathrm{TSC}_i = \frac{1}{T} \sum_{t=1}^{T} \mathrm{SC}_i(t)
\label{eq:tsc}
\end{equation}
Group- and population-level temporal self-consistency are obtained by averaging $\mathrm{TSC}_i$ across individuals within protected groups and the full cohort, respectively. To quantify stability degradation, we further compute $\Delta \mathrm{TSC}_i = \mathrm{SC}_i(1) - \mathrm{SC}_i(T)$.


\textbf{Prediction Flip Rate} captures local prediction changes between consecutive retraining phases. Let $\hat{y}_i(t) \in \{0,1\}$ denote the prediction for individual $i$ at retraining phase $t$, and let $T$ be the total number of retraining phases. A prediction flip occurs when:
\begin{equation}
\hat{y}_i(t) \neq \hat{y}_i(t+1)
\label{eq:fr_cond}
\end{equation}

The corresponding flip indicator is defined as:
\begin{equation}
\mathrm{Flip}_i(t) =
\begin{cases}
1, & \text{if } \hat{y}_i(t) \neq \hat{y}_i(t+1), \\
0, & \text{otherwise}
\end{cases}
\label{eq:fr_def}
\end{equation}

The population-level flip rate at phase $t$ is computed as:
\begin{equation}
\mathrm{FlipRate}(t) = \frac{1}{N} \sum_{i=1}^{N} \mathrm{Flip}_i(t),
\label{eq:frate}
\end{equation}
where $N$ denotes the number of individuals in the evaluation cohort.

To characterize patient-level instability during retraining, we operationalized instability using complementary criteria based on prediction flip rate and temporal self-consistency.
First, if more than 20\% of the predictions for an individual flip across consecutive retraining phases, that individual is counted as one for which predictions are unstable, otherwise, the individual is counted as one for which predictions are stable.
Given binary predictions, and in our setting in which there are five time-steps, this threshold corresponds to at least one prediction flip over the retraining experiment. 

Second, temporal self-consistency was analyzed using two complementary indicators of stability drift. An individual is counted as one for which predictions are unstable if (i) their self-consistency score exhibited a monotonic or overall decreasing trend across retraining phases, suggesting progressive degradation of prediction stability, or (ii) their minimum self-consistency value across retraining fell below 0.75, indicating periods of pronounced instability. Otherwise, the individual is counted as one for which predictions are stable.


\subsubsection{Model Multiplicity}
We assess model multiplicity through Rashomon set analysis~\cite{dai2025intentional}, identifying families of models with near-optimal performance.
Let $\mathcal{R}_t = \{f^{(1)}_t,\dots,f^{(M)}_t\}$ denote the Rashomon set at retraining phase $t$. Here, the multiplicity metrics are characterized via:

\textbf{Distinct predictive patterns (DPR)} within the Rashomon set defined as the vector of predictions assigned by a model to all individuals:
\begin{equation}
\mathrm{DPR}(t) = \left| \left\{ (\hat{y}^{(m)}_1(t), \dots, \hat{y}^{(m)}_N(t)) : f^{(m)}_t \in \mathcal{R}_t \right\} \right|
\label{eq:dpr}
\end{equation}
\textbf{Prediction Disagreement Rate (DR)} defined as the expected pairwise prediction mismatch within the Rashomon set.
\begin{equation}
\mathrm{DR}(t) =
\frac{2}{M(M-1)} 
\sum_{m<m'}
\frac{1}{N} \sum_{i=1}^{N}
\mathbb{I}[\hat{y}^{(m)}_i(t) \neq \hat{y}^{(m')}_i(t)]
\label{eq:dr}
\end{equation}


\subsection{Uncertainty-Aware Abstention for Safe Updates}


To enhance safety during deployment, we introduce an abstention mechanism 
in which this AI/ML decision support system does not issue a prediction/recommendation about uncertain cases.
We consider that abstained samples introduce additional costs, given that in those cases clinicians do not receive the benefit of having a prediction to help them make a decision.

\subsubsection{Conformal Predictions Framework}
Abstention is based in a distance-based conformal prediction framework~\cite{angelopoulos2021gentle}.
In this framework, any test sample that is not similar enough to any training sample is considered an out-of-distribution sample for which predictions are not reliable.
Specifically, for each test sample $x$, we compute the average of the distances to its $k$-nearest-neighbor in the training set:
\begin{equation}
d(x) = \frac{1}{k} \sum_{j=1}^{k} \| x - x_j^{\text{train}} \|.
\label{eq:dpr}
\end{equation}
Predictions having $d(x)$ above a percentile-based threshold $\tau$ are abstained, corresponding to a fixed abstention budget (e.g., maximum of 5\% abstention rate).
The goal is to avoid compromising overall system utility.

\subsubsection{Metrics}
We log abstention frequency, protected group membership, and prediction scores.
Performance and algorithmic fairness metrics are recomputed without the abstained samples, to assess whether abstention has an impact on model reliability, temporal stability, or fairness.

\section{Results}

\subsection{Experimental Data and Prediction Task}
We evaluated the proposed framework on four publicly available U.S.-based Type~1 Diabetes (T1D) datasets containing high-resolution continuous glucose monitoring (CGM) data.
The datasets ($\mathrm{DB}_i$, $i = 1,\dots,4$) comprise approximately 11{,}300 weekly observations from a total of 496 participants (range: 99--181 individuals per dataset, and 2--57 weeks of observations per individual).
Participants across all datasets are under 20 years of age (e.g. pediatric medical care typical age threshold; including toddlers, children, adolescents and young adults), most of them are children.
Additionally, each dataset contains structured sociodemographic information, including age, sex, and race/ethnicity.

All raw datasets were integrated and processed using a structured preprocessing pipeline to ensure consistency and clinical interpretability across sources. 
Sociodemographic variables were standardized and encoded uniformly to enable cross-database comparability. 
Continuous glucose monitoring (CGM) time series were processed to extract clinically meaningful features at both daily and event-specific levels. Daily glycemic control metrics included time in range (TIR; 70--180~mg/dL), time above range (TAR; $>$180~mg/dL), time below range (TBR; $<$70~mg/dL), standard deviation (SD), mean amplitude of glycemic excursions (MAGE), and coefficient of variation (CV), providing a compact summary of overall glycemic control and variability.  All preprocessing was performed in Python (v3.10) using pandas, numpy, and scipy within a JupyterLab environment.
In addition to daily summaries, CGM traces were segmented into hyperglycemic and hypoglycemic events by grouping consecutive glucose measurements exceeding predefined clinical thresholds.
%
%
All features were aggregated into clinically relevant temporal windows to support weekly-level prediction and discriminatory assessment.

The prediction task was formulated as a binary classification problem at the weekly level. 
Each patient-week was labeled as either \emph{low risk} or \emph{high risk} for severe hyperglycemia. 
High-risk weeks were defined as those with more than three severe hyperglycemic events, where an event was operationalized as glucose levels exceeding 250~mg/dL for a duration of at least 180~minutes, consistent with established clinical CGM guidelines\cite{battelino2019clinical,maiorino2020effects,mouri2023hyperglycemia,care2023standards}.

Each experimental dataset was divided into six chronological batches for continual retraining simulations, with patients uniquely assigned to a single batch to prevent overlap across retraining phases.
Batches were created by first ordering each dataset by observation date and patient, then assigning patients to retraining batches based on the date of their first recorded observation, ensuring that each batch represents approximately equal time spans and that patients remain in a single batch.
Disparity analyses were conducted separately for four binarized protected attributes: sex, age, caregiver educational level, and caregiver annual income, allowing the systematic evaluation of performance, fairness, and stability across clinically and sociodemographically relevant subgroups.

\subsection{Performance, Fairness, and Stability Under Model Updates}
Across both prospective and retrospective evaluation settings, we observed consistent differences among retraining strategies in terms of predictive performance, fairness, and stability. 
In terms of performance, full cumulative retraining and random subsampling of past data consistently achieved the highest and most stable AUC values, with comparable performance between the two strategies. 
Both approaches outperformed last-batch retraining and no retraining, although we note that no retraining was not uniformly the worst-performing strategy across all datasets and time periods. 
In contrast, last-batch retraining exhibited larger performance fluctuations and was associated with increased disparities between protected subgroups (see Table~\ref{tab:strategy_fairness}).


The analysis of algorithmic fairness metrics revealed a systematic disadvantage of last-batch retraining. 
Specifically, this strategy consistently amplified subgroup disparities under both DP and EO, indicating that relying solely on the most recent data batch may exacerbate inequities during continual model updates. 
By comparison, full and subsampling strategies yielded lower and more stable fairness disparities over time, although in some datasets a trade-off between predictive performance and fairness was observed, highlighting the need for multi-objective evaluation beyond accuracy alone.

\begin{figure}[b]
    \centering
    \includegraphics[width=\columnwidth]{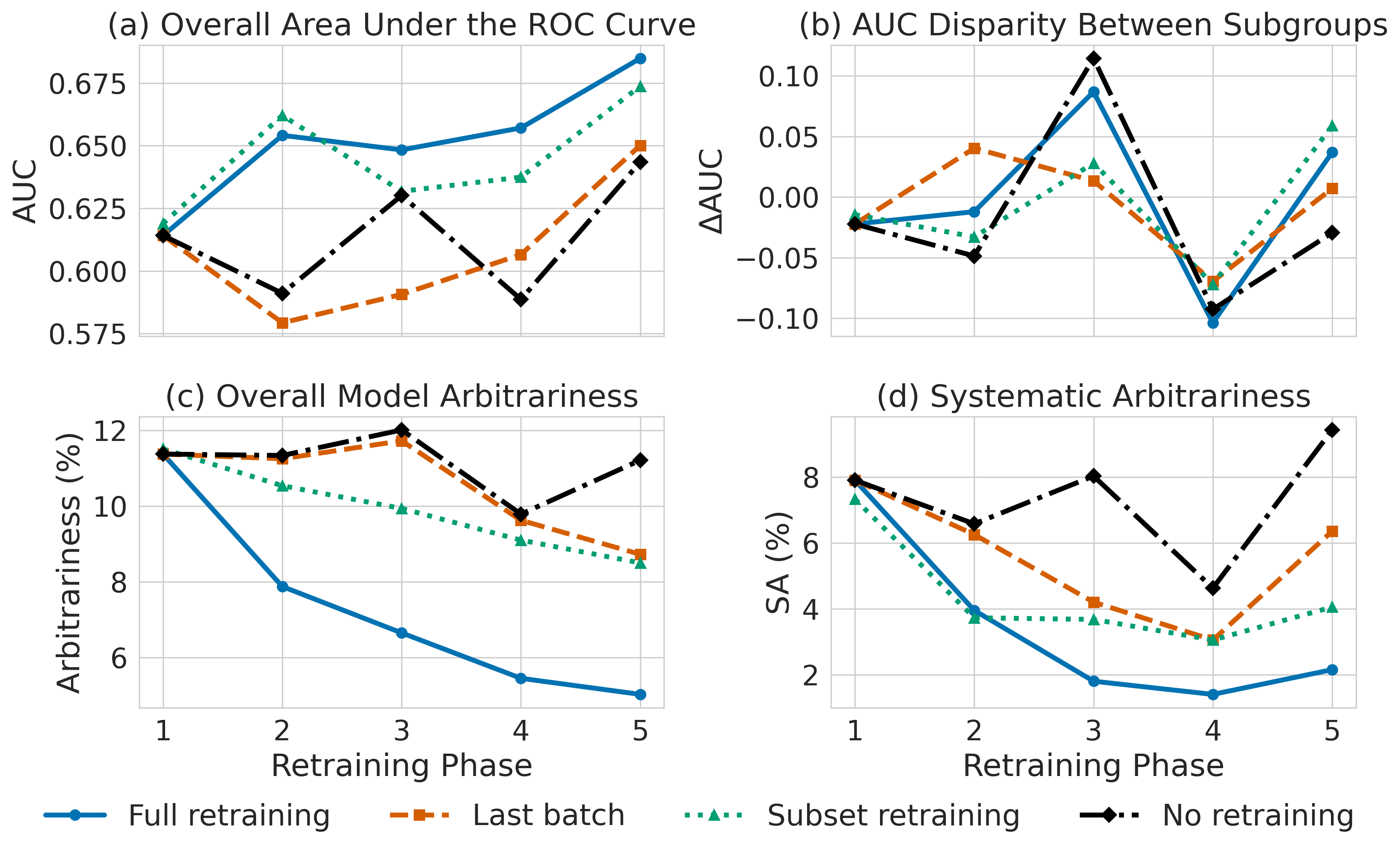}
    \caption{\textbf{Temporal evolution of aggregated average predictive performance, fairness, and stability under different continual retraining strategies across databases with Sex as protected feature}. Panels show: (a) overall area under the ROC curve, (b) disparity in area under the ROC curve between protected subgroups, (c) overall model arbitrariness, and (d) systematic arbitrariness between subgroups, across retraining phases. Lower values in fairness and arbitrariness metrics indicate improved equity and stability.}
    \label{fig:retraining_comparison}
\end{figure}

From a stability perspective, full cumulative retraining demonstrated the highest overall stability, with lower levels of systematic arbitrariness across retraining phases.
In contrast, last-batch retraining resulted in markedly higher arbitrariness, particularly manifested as stability disparities between protected subgroups. 
The findings summarized in Fig.~\ref{fig:retraining_comparison} indicate that strategies optimized for short-term adaptation may introduce instability and inequity in longitudinal clinical deployment. 

In addition to aggregate performance, fairness, and stability, we analyzed the evolution of the Rashomon set under the full cumulative retraining strategy in both retrospective and prospective validation. 
Across datasets, the prediction disagreement rate (DR) remained within a moderate range (approximately 20–32\%), generally decreasing over successive retraining phases, indicating increasing consensus among well-performing models as more data accumulated. 
An exception was observed in $\mathrm{DB}_2$, where DR exhibited a slight upward trend, suggesting persistent uncertainty in specific regions of the feature space.
Conversely, the number of DPR ranged from 1 to 7 and tended to increase over time, reflecting growing model diversity despite stable aggregate performance. 
This trend was not uniform across datasets: in $\mathrm{DB}_3$, DPR displayed oscillatory behavior without a clear monotonic pattern. 
Overall, these results show that under full retraining, increased predictive stability and agreement can coexist with growing diversity in the Rashomon set, revealing a nuanced interplay between consensus, arbitrariness, and temporal stability.

\begin{table}[htbp]
\caption{Retraining Strategy Comparison Across Protected Features}
\label{tab:strategy_fairness}
\centering
\scriptsize
\renewcommand{\arraystretch}{1.15}
\begin{tabular}{|c|c|c|c|c|c|c|}
\hline
\textbf{Protected} & \textbf{Strategy} &
\textbf{Av. AUC} &
\textbf{$\Delta$AUC} &
\textbf{EO} &
\textbf{DP} &
\textbf{OA} \\

\hline
\text{Sex}
 & No   & 0.61 & 0.11 & 0.21 & 0.22 & 0.11  \\
 & Last & 0.61 & 0.07 & 0.13 & 0.16 & 0.11  \\
 & Sub  & 0.64 & 0.07 & 0.17 & 0.18 & 0.10  \\
 & Full & 0.65 & 0.08 & 0.16 & 0.18 & 0.07  \\
\hline
\text{Age}
 & No   & 0.62 & 0.08 & 0.16 & 0.15 & 0.11  \\
 & Last & 0.61 & 0.09 & 0.17 & 0.18 & 0.11  \\
 & Sub  & 0.65 & 0.09 & 0.13 & 0.13 & 0.10  \\
 & Full & 0.65 & 0.08 & 0.12 & 0.12 & 0.07  \\
\hline
\text{Educational Level}
 & No   & 0.59 & 0.07 & 0.25 & 0.28 & 0.10  \\
 & Last & 0.60 & 0.09 & 0.20 & 0.20 & 0.11  \\
 & Sub  & 0.64 & 0.07 & 0.15 & 0.17 & 0.09  \\
 & Full & 0.64 & 0.10 & 0.17 & 0.18 & 0.07  \\
\hline
\text{Annual Income}
 & No   & 0.60 & 0.11 & 0.19 & 0.19 & 0.11  \\
 & Last & 0.60 & 0.10 & 0.19 & 0.18 & 0.11  \\
 & Sub  & 0.64 & 0.09 & 0.17 & 0.17 & 0.10  \\
 & Full & 0.64 & 0.10 & 0.17 & 0.18 & 0.08  \\
\hline
\multicolumn{7}{p{0.95\columnwidth}}{\footnotesize
$\Delta$AUC denotes absolute subgroup AUC difference; EO and DP denote Equal Opportunity and Demographic Parity disparities; OA denotes overall arbitrariness.}
\\
\end{tabular}
\end{table}

\subsection{Temporal Stability and Subgroup-Level Disparities}
\begin{figure}[!t]
\centering
\includegraphics[width=\columnwidth]{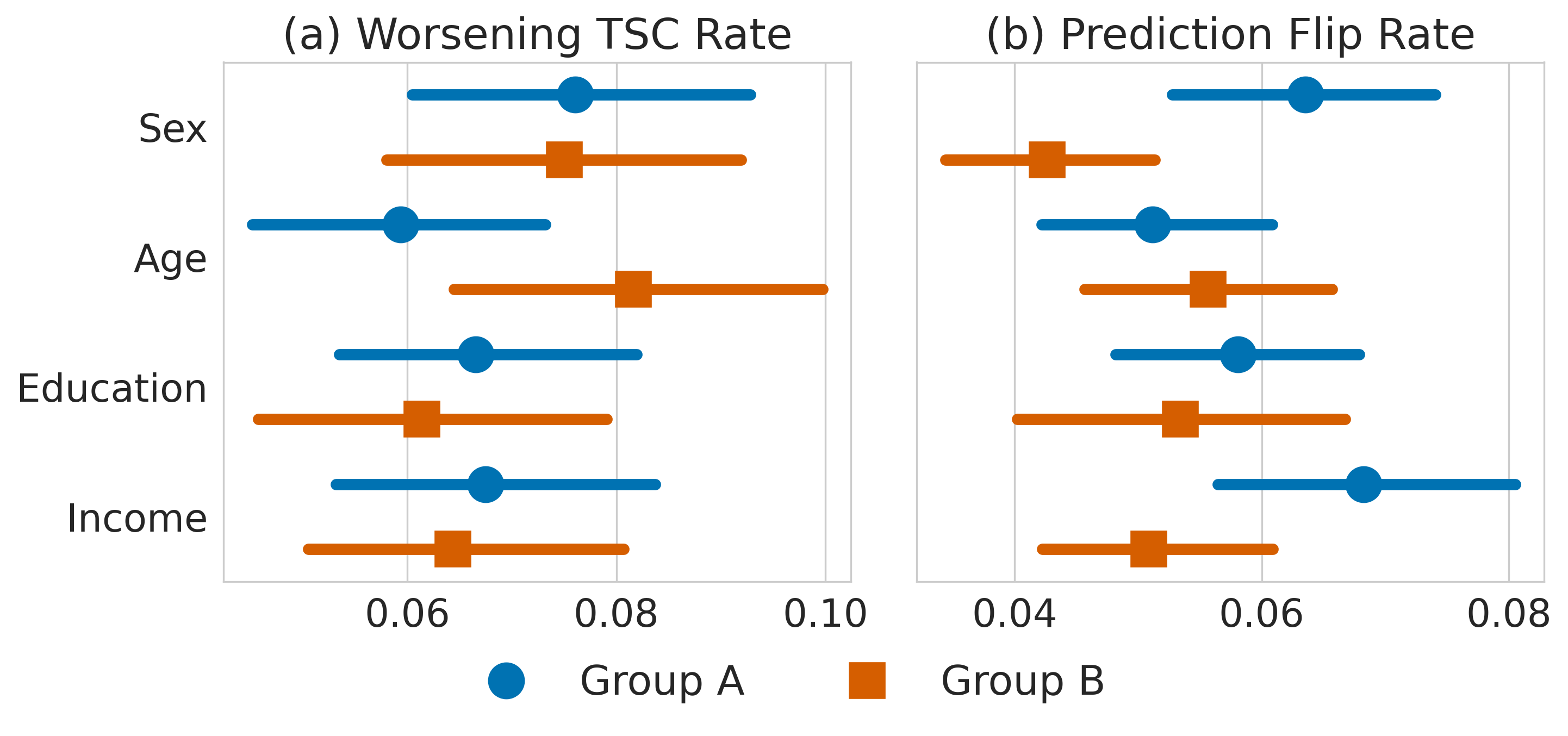} 
\caption{\textbf{Subgroup-level temporal stability analysis across protected attributes}. Panels report patient-level (a) worsening temporal self-consistency rates across retraining phases, and (b) prediction flip rates. Points denote subgroup means with 95\% confidence intervals. For each protected attribute, Group A and Group B correspond respectively to: sex (male vs. female), age (younger vs. older children), caregiver education (higher vs. lower educational level), and household income (higher vs. lower income). Older children, female children, and individuals from lower-income or lower-education households exhibit higher instability, highlighting disparities that are not captured by aggregate performance metrics.}
\label{fig:temporal_stability_fliprate}
\end{figure}

Analysis of temporal self-consistency and prediction flip rates revealed persistent subgroup-level disparities in model stability across datasets as shown in Fig.~\ref{fig:temporal_stability_fliprate}. 
Flip rates, defined as the fraction of prediction instances for which models trained in two consecutive retraining phases yield discordant outputs, range from 0\% to 7\%. 
Temporal self-consistency exhibits greater variability: the low-TSC proxy affects on average 31.9\% of individuals (range 0–100\%), indicating episodes of pronounced instability, while the worsening-TSC proxy captures progressive degradation over time, with a mean of 7\% ± 10\% (range 0–70\%).
In our setting, each individual contributes multiple weekly observations; flip rates and instability flags are first computed at the data-point (week) level and subsequently aggregated to the individual and subgroup levels, which are the quantities reported in Table~\ref{tab:temporal_instability}. 
%
In particular, predictions for older children, female children, and individuals from lower-income or lower educational level households exhibited higher flip rates and reduced temporal consistency over successive retraining phases. 
These patterns were observed consistently across multiple datasets, suggesting that instability disproportionately affects already vulnerable patient groups.

Importantly, these disparities were not always reflected in aggregate performance metrics, underscoring that temporal stability captures complementary information about model behavior that is not observable through accuracy or AUC alone. 
This further supports the need to explicitly monitor longitudinal consistency when deploying adaptive clinical models.

\begin{figure}[!b]
    \centering
    \includegraphics[width=\columnwidth]{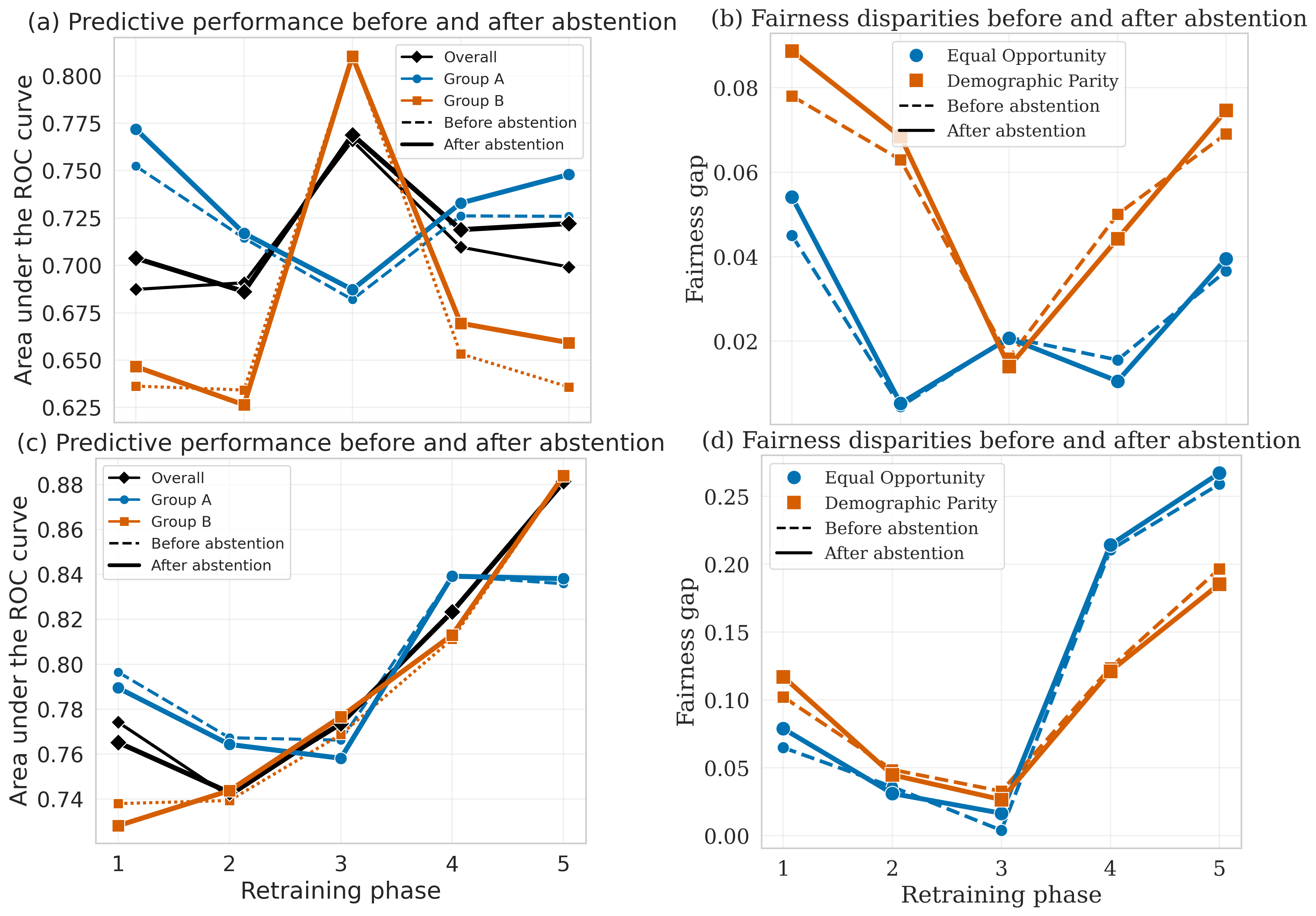}
    \caption{\textbf{Representative examples illustrating heterogeneous effects of abstention across datasets and protected features}.
    Panels (a)–(b) correspond to DB4 with \textit{household income} as the protected feature, where abstention leads to modest improvements in overall predictive performance (AUC) as well as reductions in both Equal Opportunity (EO) and Demographic Parity (DP) disparities.
    In contrast, panels (c)–(d) show DB1 with \textit{age} as the protected feature, where abstention has negligible impact on predictive performance, while inducing divergent fairness effects: EO disparity increases whereas DP disparity improves.}
    \label{fig:abstention_perf_fairness_examples}
\end{figure}

\subsection{Impact of Distance-Based Conformal Abstention}
The proposed distance-based conformal prediction (CP) abstention mechanism exhibited heterogeneous effects across datasets and subgroups.
While abstention generally improved overall predictive performance and reduced instability on retained predictions, performance gains and fairness improvements were not uniformly observed across all protected subgroups as shown in Fig.~\ref{fig:abstention_perf_fairness_examples}(a)--(b). 
In some cases, subgroup-level AUC did not increase after abstention, and fairness disparities under Demographic Parity or Equal Opportunity were only partially mitigated (see Fig.~\ref{fig:abstention_perf_fairness_examples}(c)--(d)).

We further observed that abstention disproportionately affected patients from lower educational level and lower income households, as well as male children, indicating that uncertainty-aware abstention may itself introduce differential impacts across population strata (see Table~\ref{tab:temporal_instability}). 
These findings highlight the importance of carefully auditing abstention mechanisms for unintended equity effects.

To better understand the nature of abstained samples, we analyzed their temporal stability characteristics.
In several datasets, most notably $\mathrm{DB}_1$, a substantial proportion of abstained patients exhibited substantially higher prediction flip rates and lower temporal self-consistency compared to retained patients. 
%
This suggests that the abstention mechanism effectively identifies unstable or out-of-distribution predictions. Importantly, these flagged predictions are not acted upon by the system; they are exclusively routed for clinician review, ensuring that all final interpretations remain under human supervision.

\section{Discussion and Conclusion}

Our study shows that failing to update artificial intelligence and machine learning models in clinical settings can reduce predictive performance, increase subgroup disparities, and degrade temporal stability. 
While model updates are necessary, they may also introduce unintended effects that require continuous monitoring and auditing. Using pediatric Type 1 Diabetes prediction as a case study, we proposed and evaluated a framework for assessing the impact of retraining strategies on performance, fairness, arbitrariness, and longitudinal stability.

The appropriate update strategy and retraining frequency should be determined empirically for each deployment context. 
Although our findings are based on representative pediatric diabetes datasets and may not generalize to all clinical settings, they highlight the importance of monitoring adaptive clinical models beyond aggregate accuracy metrics alone.

\begin{table}[!t]
\caption{Prevalence of Temporal Instability Across Protected Subgroups}
\label{tab:temporal_instability}
\centering
\scriptsize
\renewcommand{\arraystretch}{1.1}
\begin{tabular}{|c|c|c|c|c|}
\hline
\makecell{\textbf{Protected} \\ \textbf{Feature}} &
\textbf{Subgroup} &
\makecell{\textbf{\% Instability} \\ \textbf{(FR)}} &
\makecell{\textbf{\% Instability} \\ \textbf{(Low TSC)}} &
\makecell{\textbf{\% Abstention} \\ \textbf{High Rate}} \\
\hline
\multirow{2}{*}{Age}
 & Older   & 4.22 & 11.45 & 4.84 \\
 & Younger & 1.29 & 9.68 & 7.85 \\
\hline
\multirow{2}{*}{Sex}
 & Female & 5.16 & 12.90 & 6.77 \\
 & Male   & 4.22 & 9.70 & 5.93\\
\hline
\multirow{2}{*}{Income}
 & Lower  & 5.38 & 16.92 & 8.41 \\
 & Higher & 3.59 & 13.17 & 5.28 \\
\hline
\multirow{2}{*}{Education}
 & Lower  & 5.11 & 9.09 & 7.99\\
 & Higher & 4.44 & 8.89 & 5.49\\
\hline
\multicolumn{5}{p{0.9\columnwidth}}{\footnotesize
Unstable individuals are defined as having a prediction flip rate greater than 20\% across retraining phases or exhibiting temporal self-consistency below 0.75 at any point during continual retraining. High abstention rate is the fraction of individuals with more than 10\% abstained weeks. Values are averaged across datasets.}
\\
\end{tabular}
\end{table}
Future work should extend this approach to other clinical settings, incorporate additional metrics of model quality, and explore retraining strategies informed by population- and feature-specific drifts.

\section*{Ethics Statement and Data Availability}
\label{data_avail}
This study used de-identified publicly available pediatric Type 1 Diabetes datasets, previously collected under clinical trial protocols approved by the corresponding Institutional Review Boards and ethics committees, coordinated by the Jaeb Center for Health Research (JAEB), including PEDAP \cite{PEDAP_dataset}, IOBP2 \cite{IOBP2_dataset}, DCLP5 \cite{DCLP5_dataset}, and CITY \cite{CITY_dataset}, accessed under applicable data use terms, with no new data collection or participant interaction.


\bibliographystyle{IEEEtran}
\bibliography{ref}



\end{document}